\documentclass[sn-mathphys-num]{sn-jnl}

\RequirePackage{amsthm}
\usepackage{lscape}
\usepackage{changepage}
\usepackage{tabularx} 
\usepackage{lipsum} 
\usepackage{adjustbox} 
\usepackage{balance}
\usepackage{graphicx}
\usepackage{hyperref}
\usepackage{lscape}
\usepackage{longtable}
\usepackage{graphicx}%
\usepackage{multirow}%
\usepackage{amsmath,amssymb,amsfonts}%
\usepackage{amsthm}%
\usepackage{mathrsfs}%
\usepackage[title]{appendix}%
\usepackage{xcolor}%
\usepackage{textcomp}%
\usepackage{manyfoot}%
\usepackage{booktabs}%
\usepackage{algorithm}%
\usepackage{algorithmicx}%
\usepackage{algpseudocode}%
\usepackage{listings}%
\usepackage{tabularx}%
\usepackage[utf8]{inputenc}%
\UseRawInputEncoding
\usepackage{mathrsfs,amsmath}

\theoremstyle{thmstyleone}%
\theoremstyle{thmstyletwo}%

\theoremstyle{thmstylethree}%

\begin{document}

\title[Article Title]{Multi-scale Frequency Enhancement Network for Blind Image Deblurring}

\author[1]{\fnm{Yawen} \sur{Xiang}} 

\author[2,4]{\fnm{Heng} \sur{Zhou}} 

\author[3,5]{\fnm{Chengyang} \sur{Li}} 

\author*[1,]{\fnm{Zhongbo} \sur{Li}}\email{lzb05296@163.com}
\author*[1,]{\fnm{Yongqiang} \sur{Xie}}\email{yqxie$\_$2024@163.com}

\affil[1]{\orgdiv{The Institute of Systems Engineering}, \orgname{Academy of Military Science}, \city{Beijing}, \postcode{100141},  \country{China}}
\affil[2]{\orgdiv{School of Artificial Intelligence and Computer Science}, \orgname{Jiangnan University}, \city{Wuxi }, \postcode{214122},  \country{China}}
\affil[3]{\orgdiv{College of Artificial Intelligence}, \orgname{China University of Petroleum(beijing)}, \city{Beijing}, \postcode{100100},  \country{China}}
\affil[4]{\orgdiv{Jiangsu Provincial Engineering Laboratory of Pattern Recognition and Computational Intelligence}, \orgname{Jiangnan University}, \city{Wuxi}, \postcode{214122},  \country{China}}
\affil[5]{\orgdiv{Beijing Key Laboratory of Petroleum Data Mining}, \orgname{China University of Petroleum(beijing)}, \city{Beijing}, \postcode{100100},  \country{China}}


\abstract{ Image deblurring is an essential image preprocessing technique, aiming to recover clear and detailed images form blurry ones. However, existing algorithms often fail to effectively integrate multi-scale feature extraction with frequency enhancement, limiting their ability to reconstruct fine 	textures. Additionally, non-uniform blur in images also restricts the effectiveness of image restoration. To address these issues, we propose a multi-scale frequency enhancement network (MFENet) for blind image deblurring. To capture the multi-scale spatial and	channel information of blurred images, we introduce a multi-scale feature extraction module (MS-FE) based on depthwise separable convolutions, which provides rich target features for deblurring. We propose a frequency enhanced blur perception module (FEBP) that employs wavelet transforms to extract high-frequency details and utilizes multi-strip pooling to perceive non-uniform blur, combining multi-scale information with frequency enhancement to improve the restoration of 	image texture details. Experimental results on the GoPro and HIDE datasets demonstrate that the proposed method achieves superior deblurring performance in both visual quality and objective evaluation metrics. Furthermore, in downstream object detection tasks, the proposed blind image deblurring algorithm significantly improves detection accuracy, further validating its effectiveness androbustness in the field of image deblurring.

}




\maketitle

\section{Introduction}\label{sec1}

With the increasing use of sensors and imaging devices, images have become an important medium for information. However, during the image acquisition process, motion blur inevitably occurs due to rapid object movement or camera shake. Blurred images not only negatively affect visual perception but also have adverse effects on downstream computer version tasks such as image segmentation and object detection.

Early methods mainly focused on non-blind deblurring algorithms{~\cite{arjomand2017deep,sun2014good,zhang2017learning}, which utilized prior knowledge, such as sparse prior~\cite{krishnan2011blind}, patch prior~\cite{sun2013edge}, and gradient prior~\cite{pan2014deblurring}, to estimate the blur kernel. Using the estimated blur kernel $\mathit{k}$, deconvolution operations are performed on the blurred images through inverse filtering, Wiener filtering, R-L filtering, and regularization algorithms to obtain sharp images. These deblurring methods rely heavily on the accurate estimation of the blur kernel $\mathit{k}$. However, in real-world scenarios, both the blur kernel $\mathit{k}$ and the noise $\mathit{n}$ in the equation are unknown,  making the inverse problem of solving for the sharp image $\mathit{I}$ an ill-posed problem with no unique solution. Consequently, non-blind deblurring algorithms have significant limitations in practical applications and are often inadequate for handling complex blur in real-world scenes.
	
	In recent years, many deep learning-based methods~\cite{zhao2022rethinking,purohit2021spatially,chen2022simple} have been proposed. These methods learn the nonlinear mapping relationship between the blurred image and the sharp image through end-to-end training\cite{Xiang2024deep}. They are better equipped to handle non-uniform blur in real-world scenarios. Aiming at the problem that most methods fail to integrate multi-scale feature extraction with frequency enhancement and insufficiently consider the issue of non-uniform blur, which restrict their ability to reconstruct fine textures, we propose a novel image deblurring method based on multi-scale blur perception and frequency enhancement. This method employs a single U-Net as the backbone network, combining multi-scale feature extraction with frequency enhancement, thereby integrating multi-scale features extracted from different network layers to achieve improved deblurring performance. The multi-scale feature extraction module (MS-FE) effectively captures multi-scale spatial and channel features of the images, facilitating a better understanding of both the overall structure and local details, which enhances the capability to handle non-uniform blur. The frequency enhanced blur perception module (FEBP) can simultaneously handle spatial and frequency domain features, offering superior processing capabilities for non-uniform blur in complex scenes. By employing multi-strip pooling, the network is able to perceive blurred regions in both horizontal and vertical directions, enabling targeted enhancement of deblurring performance based on its orientation. Furthermore, the use of discrete wavelet transform (DWT)~\cite{zhang2024image} allows for the decomposition of the image into distinct frequency components, converting spatial domain features into the wavelet domain. This method can effectively restore edge and detail information in the image and thereby reducing detail loss due to blurring. The main contributions of this paper are summarized as follows.

	\begin{itemize}
		\item We design a multi-scale feature extraction module (MS-FE) using depthwise separable convolution to capture details at various levels within the image, thereby enhancing	the understanding of its overall structure and local details.

		\item We design a frequency enhanced blur perception module (FEBP) that utilizes multi-strip pooling to perceive non-uniform blur in images. Additionally, we employ wavelet transform to capture frequency-domain information, supplementing the texture details of the image and enhancing the recovery of image textures.
		
		\item Extensive experiments have demonstrated that our MFENet achieves better visual effect and quantitative metrics than the state-of-the-art techniques on the GoPro and HIDE datasets. Futhermore, in downstream object detection tasks, the proposed blind image deblurring algorithm significantly improves detection accuracy.

	\end{itemize}

	In the subsequent sections of this paper, section \ref{sec2} of this paper introduces the related research work of deep learning methods for image deblurring. Section \ref{sec3} details the proposed methods and network structure. Section \ref{sec4} describes the experimental details and analyzes the experimental results. Section \ref{sec5} summarizes the full text.

	\section{Related work}\label{sec2}

	Image deblurring in complex scenes presents a significant challenge, as the blur within these scenarios is often non-uniform. Different regions within an image may exhibit varying degrees, directions, and extents of blur. With the advancement of deep learning, the application of deep neural networks in image deblurring has become increasingly prevalent.
	Deep learning-based methods learn the nonlinear mapping from blurred images to sharp images in an end-to-end manner, requiring minimal prior knowledge to achieve satisfactory deblurring performance.

	Xu et al.~\cite{xu2014deep} were the first to use CNNs to achieve end-to-end restoration of blurred images. To better handle spatially varying blur, Nah et al.\cite{nah2017deep} proposed a multi-scale image deblurring algorithm based on a CNN network to remove blur in complex scenes. This method uses multi-scale images as input and progressively restores sharp images from coarse to fine using multi-scale image information. Tao et al.~\cite{tao2018scale} proposed a scale-recurrent deblurring network that enhances image restoration while reducing computational complexity through cross-scale weight sharing. To address the inefficiency of cross-scale information propagation and the loss of information during downsampling in previous multi-scale structures, Kim et al.\cite{kim2022mssnet} proposed a multi-scale phase network for image deblurring. Waqas et al.~\cite{zamir2022learning} introduced a method that encodes multi-scale context through parallel convolutional streams, restoring sharp images from coarse to fine. Zamir et al.~\cite{zamir2021multi} proposed a multi-stage progressive image restoration network that enhances the effectiveness of image deblurring using multi-scale contextual information. Chen et al.~\cite{chen2021hinet} designed a multi-level network called HINet, which enhances image restoration by fusing information from different levels across stages.

	As important contour information is lost during the blurring process, more researchers have recognized the significance of frequency information. Zou et al.~\cite{zou2021sdwnet} proposed an end-to-end CNN model in the wavelet domain, utilizing the frequency domain characteristics of wavelet transforms to provide additional information for spatial reconstruction. This approach supplements the spatial domain with information recovered in the frequency domain, resulting in restored images with more high-frequency details. Recognizing the critical role of the frequency domain in image restoration, Zhang et al.\cite{zhang2024image} embedded wavelet transforms into a deep residual network. This method converts spatial domain features into the wavelet domain, fully utilizing the texture structure information in the high-frequency sub-bands to restore edge contours and other detailed features of blurred images. Kong et al.~\cite{kong2023efficient} explored the characteristics of Transformers in the frequency domain and developed a frequency-domain-based self-attention solver. This approach effectively utilizes useful frequency information, resulting in deblurred images with sharp structural details. 
	Dong et al.~\cite{dong2023multi} proposed a simple yet effective multi-scale residual low-pass filtering network. This network can adaptively learn the spatial variation characteristics of features, effectively modeling both high-frequency and low-frequency information. To address the neglect of frequency differentiation in multi-scale methods, Zhang et al.~\cite{zhang2023multi} proposed a multi-scale frequency separation network, which combines multi-scale strategies with frequency separation to enhance the recovery of details in blurred images.
	
	In complex scenarios, the aforementioned methods often fail to address both non-uniform blur and frequency information loss simultaneously, resulting in suboptimal performance in recovering from non-uniform blur. To address these issues, we propose a multi-scale frequency enhancement network (MFENet) for blind image deblurring. MFENet integrates multi-scale feature extraction, frequency information enhancement, and blur perception. By leveraging wavelet transforms and blur attention mechanisms, MFENet improves the effectiveness of image deblurring.

	\begin{figure*}[!t]
		\centerline{\includegraphics[width=\textwidth]{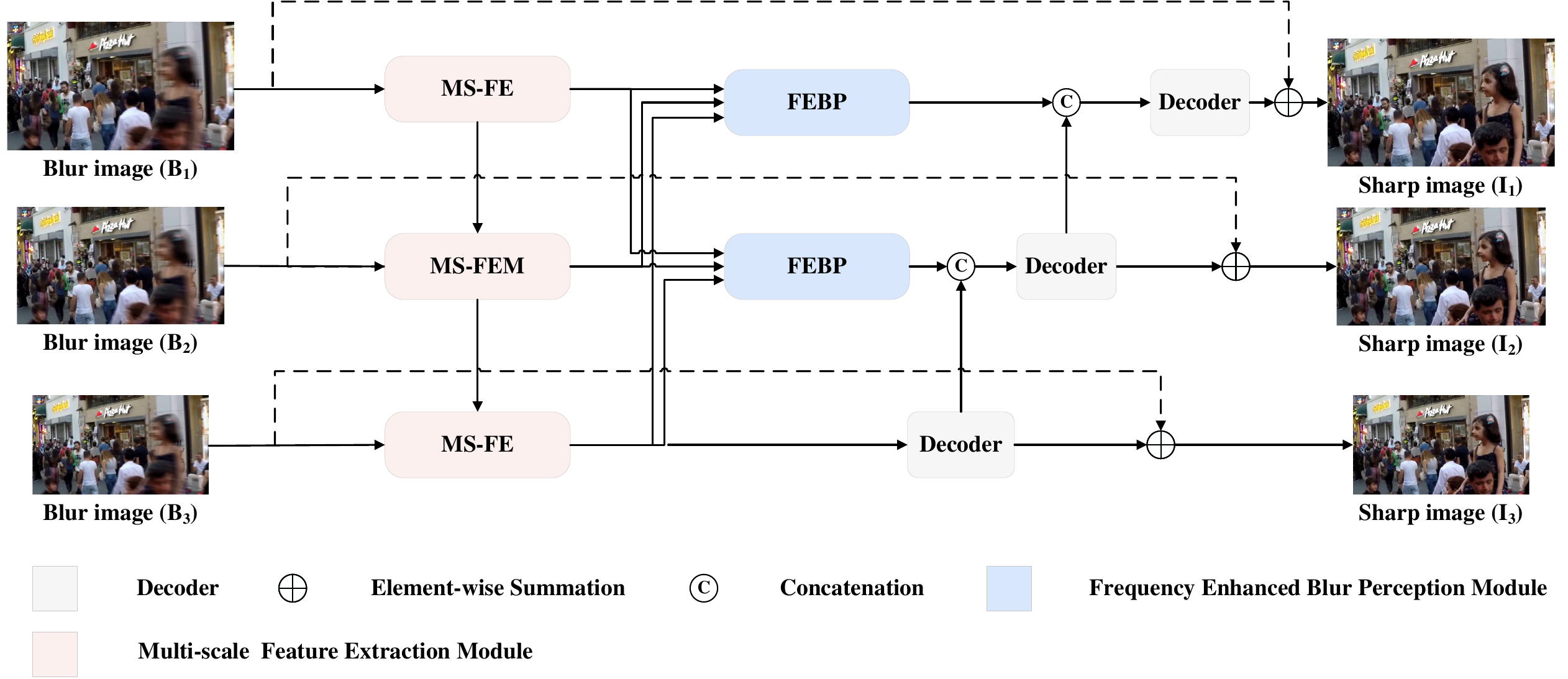}}
		\caption{The overall architecture of the proposed MFENet.\label{network}}
	\end{figure*}

	\section{Method}\label{sec3}

	Based on the multi-input multi-output U-Net~\cite{cho2021rethinking} architecture, we propose a deblurring network MFENet capable of effectively extracting multi-scale features while enhancing image frequency and blur perception. An overview of the MFENet is shown in Figure \ref{network}.  To efficiently handle non-uniform blur in images and improve the recovery of high-frequency information, the network is consisted three parts: multi-scale feature extraction module (MS-FE), frequency enhanced blur perception module (FEBP) and decoder. In this section, we will introduce the multi-scale feature extraction module (MS-FE) and the frequency enhanced blur perception module (FEBP) in detail.

	\subsection{Network Architecture}

	The architecture of MFENet is illustrated in Figure \ref{network}. The network first down-samples the input blurred image ${{B}_{1}}$ by factors of 2 and 4 to obtain two additional scaled input images ${{B}_{2}}$ and ${{B}_{3}}$. The three scales of blurred images ${{B}_{1}}$, ${{B}_{2}}$, and ${{B}_{3}}$ are processed through MS-FE to extract features. In the module, the features extracted from  ${{B}_{2}}$, and ${{B}_{3}}$ are fused with the down-sampled features extracted by the upper-level feature extractor. 
	
	In complex scenes, blur is often non-uniform, and simply extracting multi-scale blur features is insufficient to distinguish regions of varying blur levels within the images. Additionally, there is the issue of high-frequency information loss. Therefore, we introduce a FEBP module to process the fused features of three levels, assigning appropriate weights to different blurred regions to effectively implement blur perception and frequency enhancement.
	
	The decoder transfers the lower-scale features to the upper layers via transposed convolution, enabling the upper-layer decoder to utilize multi-scale image features. Then, the features are combined using skip connections to increase the information flow, fusing shallow and deep features of the network.

	\subsection{Multi-Scale Feature Extraction Module}

	To extract multi-scale information and expand the receptive field of CNN convolutions, we design a multi-scale feature extraction module (MS-FE) based on depthwise separable convolutions, as shown in Figure \ref{MS_FEM}. MS-FE consists of three main components: multi-scale depthwise convolutions , pointwise convolutions and feature fusion. Depthwise convolutions are used to extract spatial features from the image, while pointwise convolutions are used to capture channel-wise features. Feature fusion is used to fuse upper-layer information.

	In depthwise convolutions, each convolution kernel operates separately on each input channel to capture spatial features of each channel. In the depthwise convolution branch, the input blurred image ${{B}_{k}}$ first undergoes a $3\times 3$ convolution followed by a non-linear activation function, LeakyReLU, transforming it into a feature map ${{F}_{k}}$. This feature map is then input into the multi-scale depthwise convolution to extract multi-scale features as in Eq.~\eqref{FK}:
	
	\begin{equation}
		{{F}_{k}}=Leaky\operatorname{Re}LU\left( Con{{v}_{3\times 3}}\left( {{B}_{k}} \right) \right)\label{FK}
	\end{equation}

	To increase the receptive field and extract multi-scale features, four different convolution kernels are used for the preliminary feature extraction, specifically sizes of  $1\times 1$,  $3\times 3$,  $5\times 5$, and  $7\times 7$. Larger convolution kernels capture a larger receptive field, but kernels larger than $7\times 7$ are avoided to prevent excessive parameterization and reduced computational efficiency ~\cite{bao2020depthwise}. The multi-scale depthwise convolution features ${{D}_{k}}$ are obtained by fusing the feature maps processed with different-sized convolution kernels:
	\begin{equation}
		{{D}_{k}}=Concate\left( {{F}_{k}}\otimes {{K}_{Dept{{h}_{i}}}} \right)\text{    }i=1,\text{ }3,\text{ }5,\text{ }7\label{DK}
	\end{equation}
	${{K}_{Dept{{h}_{i}}}}$ represents the depthwise convolution kernel with size $i\times i$ in Eq.~\eqref{DK}.
	
	Pointwise convolutions apply weighting in the channel direction to the features obtained from the previous step. They fuse the four independent depthwise convolution features along the channel dimension. Pointwise convolution involves performing $in\_channel\times 4$ number of $1\times 1$ convolutions on the fused depthwise convolution features to capture the features at each point. The output of the pointwise convolution is denoted as ${{P}_{k}}$ in Eq.~\eqref{PK}:
	
	\begin{equation}
		{{P}_{k}}=BN\left( BN\left( {{D}_{k}} \right)\otimes {{K}_{point}} \right)\label{PK}
	\end{equation}
	where ${{K}_{point}}$ represents the convolution kernel for pointwise convolution, and ${BN}$ represents batch normalization. Finally, the features obtained from the pointwise convolution are concatenated with the input $B_{k}$, and an additional $1\times 1$ convolution layer is used to further optimize the concatenated features. The resulting multi-scale spatial and channel fusion features $M_{k}^{out}$ are given by Eq.~\eqref{MK}:
	\begin{equation}
		M_{k}=Con{{v}_{1\times 1}}\left( Concate\left( {{P}_{k}},{{B}_{k}} \right) \right)\label{MK}
	\end{equation}

	\begin{figure*}[!t]
		\centerline{\includegraphics[width=\textwidth]{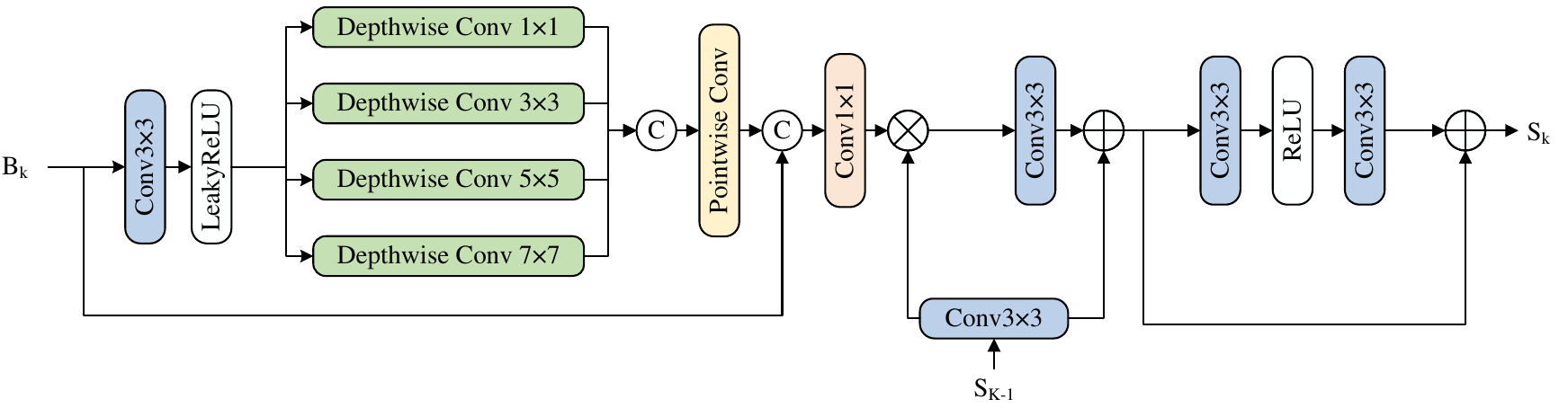}}
		\caption{Multi-scale feature extraction module.\label{MS_FEM}}
	\end{figure*}

	Then, MS-FE fuses the features extracted by the upper layer, and obtains the feature output of this layer through the residual block.The module output features ${{S}_{k}}$ are represented as:
	
	\begin{equation}
		S_{k}=M_{k}^{out}\oplus{{f}_{out}}\left( M_{k}^{out}\right)\label{SK}
	\end{equation}
	
	\begin{equation}
		M_{k}^{out}=S_{k-1}\oplus{Con{{v}_{3\times 3}}}\left( {Con{{v}_{1\times 1}}}\left(	M_{k}\right) \times {Con{{v}_{3\times 3}}}\left( S_{k-1} \right)\right)\label{MOK}
	\end{equation}
	The function \({{f}_{out}}\left( \cdot \right) = \text{Conv}\left( \sigma \text{ReLU}\left( \text{Conv}\left( \cdot \right) \right) \right)\) represents a nonlinear mapping function composed of two  $3\times 3$ convolutional layers in Eq.~\eqref{SK}. \(\oplus\) represents the feature fusion concatenation operation in Eq.~\eqref{SK} and Eq.~\eqref{MOK}. MS-FE simultaneously considers both spatial and channel domains of the image, achieving separation in both domains. Additionally, by using depthwise convolutions with multiple kernel sizes, it captures more comprehensive features, thereby enhancing the model's deblurring effectiveness.

	\begin{figure*}[!t]
		\centerline{\includegraphics[width=\textwidth]{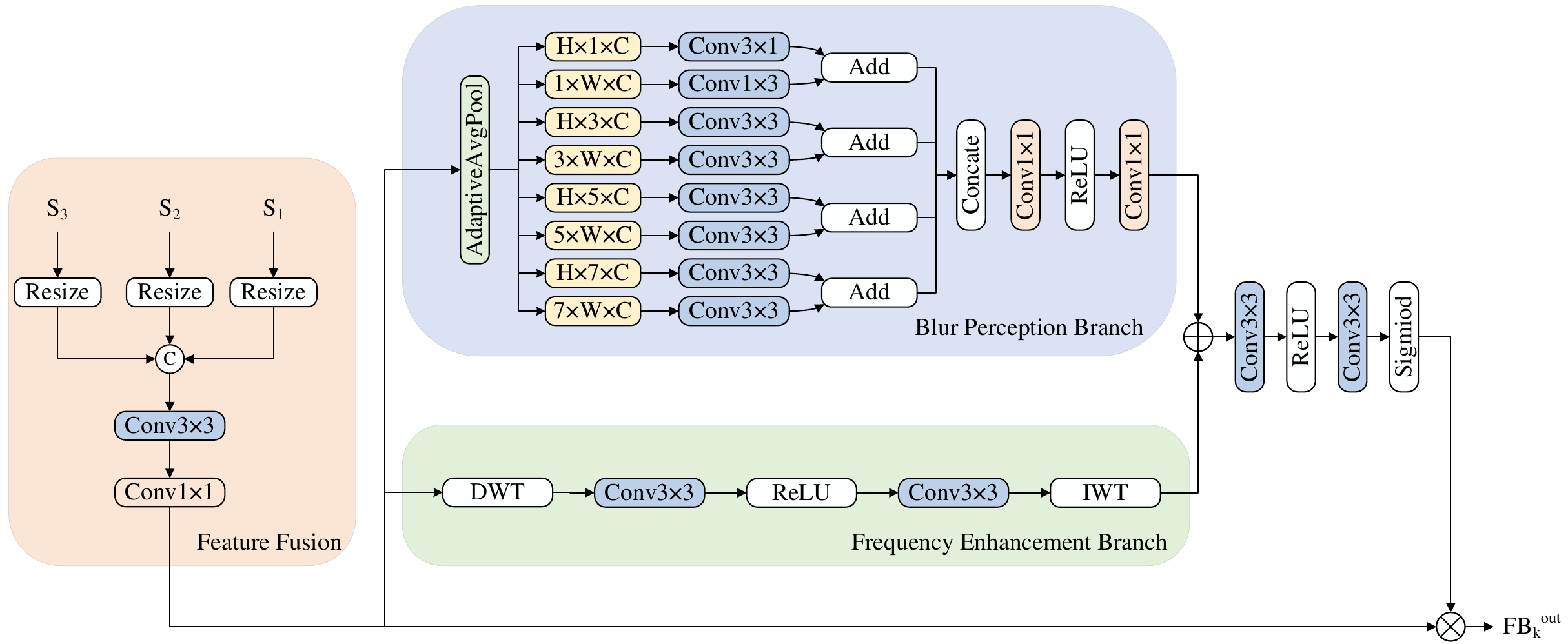}}
		\caption{Frequency enhanced blur perception module.\label{BA}}
	\end{figure*}

	\subsection{Frequency Enhanced Blur Perception Module}

	As illustrated in Figure \ref{BA}, FEBP uses different-sized and oriented strip pooling to identify region and directional averaging artifacts caused by dynamic blur~\cite{tsai2022banet}, and utilizes a wavelet transform branch to address the issue of poor texture recovery due to high-frequency information loss.
	
	FEBP first combines multi-scale features using convolution operations. In Eq.~\eqref{FM1} and Eq.~\eqref{FM2}, the results of multi-scale fusion features for the first and second layers of the network are denoted as $F{{M}_{1}}$ and $F{{M}_{2}}$:
	\begin{equation}
		F{{M}_{1}}=Con{{v}_{1\times 1}}Con{{v}_{3\times 3}}Concate(S_{1},S{_{2}^{\uparrow }},S{_{3}^{\uparrow }}) \label{FM1}
	\end{equation}
	\begin{equation}
		F{{M}_{2}}=Con{{v}_{1\times 1}}Con{{v}_{3\times 3}}Concate(S{_{1}^{\downarrow }},S_{2},S{_{3}^{\uparrow }}) \label{FM2}
	\end{equation}
	Where, $\uparrow $ represents upsampling, and $\downarrow $ represents downsampling. The fusion of multi-scale features enables the module to effectively utilize and process features at different levels. 
	
	Subsequently FEBP uses blur perception branch to capture blurs of different sizes and orientations, helping to deal with non-uniform blurs. The blur perception branch first pools the input tensors to generate the vertical band features of $H\times n\times C$ and the horizontal band features of $n\times W\times C$, where $n\in \left\{ \left. 1,\text{ 3, 5, 7} \right\} \right.$.
	The implementation of stripe pooling in both horizontal and vertical directions assists the network in capturing non-uniform blur information along the respective orientations. 
	Then, the horizontal and vertical strip features are processed by a 1D \(\left( n = 1 \right)\) or 2D \(\left( n = 3, 5, 7 \right)\) convolution layer, producing the vertical tensor \({{y}^{v,n}} \in {{R}^{H \times n \times C}}\) and the horizontal tensor \({{y}^{h,n}} \in {{R}^{n \times W \times C}}\), where the vertical tensor is:
	
	\begin{equation}
		y_{i,j,c}^{v,n}=\frac{1}{{{K}_{h}}}\sum\limits_{k=0}^{{{K}_{h}}-1}{F{{M}_{k}}_{i,\left( j\cdot {{S}_{h}}+k \right),c}} \label{yh}
	\end{equation}
	The horizontal stride is given by \({{S}_{h}} = \left\lfloor \frac{W}{n} \right\rfloor\), and the horizontal strip kernel size is ${{K}_{h}} = W - \left( n - 1 \right){{S}_{h}}$ in Eq.~\eqref{yh}. Correspondingly, in Eq.~\eqref{yn} the horizontal tensor is defined as:
	\begin{equation}
		y_{i,j,c}^{h,n}=\frac{1}{{{K}_{v}}}\sum\limits_{k=0}^{{{K}_{v}}-1}{F{{M}_{\text{k}}}_{\left( i\cdot {{S}_{v}}+k \right),j,c}}\label{yn}
	\end{equation}
	In Eq.\eqref{yn}, the vertical stride is given by \({{S}_{v}} = \left\lfloor \frac{H}{n} \right\rfloor\), and the vertical strip kernel size is \({{K}_{v}} = H - \left( n - 1 \right){{S}_{v}}\) .
	After obtaining the vertical and horizontal tensors, each pair of tensors is expanded to the size of \(H \times W\). Then, each pair of orthogonal tensors \(\left( {{y}^{v,n}},{{y}^{h,n}} \right)\) is fused into a single tensor \({{y}^{n}} \in {{R}^{H \times W \times C}}\) through feature addition as in Eq.~\eqref{yc}:
	
	\begin{equation}
		y_{i,j,c}^{n}=y_{i,\left\lfloor \frac{n\times j}{W} \right\rfloor ,c}^{v,n}+y_{\left\lfloor \frac{n\times i}{H} \right\rfloor ,j,c}^{h,n}\label{yc}
	\end{equation}

	Finally, all the fused feature tensors are concatenated to obtain the output of the blur perception branch, denoted as \({{F}_{b}}\):
	
	\begin{equation}
		{{F}_{\text{b}}}={{f}_{out}}\left( {{y}^{1}}\oplus {{y}^{3}}\oplus {{y}^{5}}\oplus {{y}^{7}} \right)
	\end{equation}
	After the feature concatenation operation, convolution is applied to the fused features to capture blur components of different sizes and directions.

	Wavelet transforms focus on frequency domain information and enable feature extraction without losing detail. FEBP introduces a frequency enhancement branch that uses the Haar wavelet transform to model frequency domain features, as shown in Figure \ref{BA}. In Eq.~\eqref{DWT}, the features input to the wavelet branch are divided into four distinct frequency sub-bands using Discrete Wavelet Transform (DWT), defined as:
	
	\begin{equation}
		{{{F}_{LL}},{{F}_{LH}},{{F}_{HL}},{{F}_{HH}}} =DWT\left( F{{M}_{k}} \right)\label{DWT}
	\end{equation}
	where \( LL \) denotes the low-frequency content of the image, \( HL \) denotes high-frequency information in the horizontal direction, \( LH \) denotes high-frequency information in the vertical direction, and \( HH \) denotes high-frequency information in the diagonal direction in Eq.~\eqref{DWT}. After obtaining the four frequency sub-bands, each is embedded into a $3\times 3$ convolution, and the frequency features are extracted using activation functions and convolution layers, represented as:
	
	\begin{equation}
		{{{F}_{LL\_out}},{{F}_{LH\_out}},{{F}_{HL\_out}},{{F}_{HH\_out}}} ={f}_{out}\left( {{{F}_{LL}},{{F}_{LH}},{{F}_{HL}},{{F}_{HH}}}\right)\label{fout}
	\end{equation}
	${{f}_{out}}\left( \cdot  \right)=Conv\left( \sigma \operatorname{Re}LU\left( Conv\left( \cdot  \right) \right) \right)$ is a $3\times 3$ convolution layer in Eq.~\eqref{fout}. Then, the inverse wavelet transform reconstructs the four sub-bands into output feature values ${{F}_{wt}}$, defined as:

	\begin{equation}
		{{F}_{_{wt}}}=IDWT\left( {{F}_{LL\_out}},{{F}_{LH\_out}},{{F}_{HL\_out}},{{F}_{HH\_out}} \right)\label{IDWT}
	\end{equation}
	$IDWT\left( \cdot  \right)$ represents the inverse discrete wavelet transform in Eq.~\eqref{IDWT}. After obtaining ${{F}_{\text{b}}}$ and the wavelet domain feature extraction branch ${{F}_{wt}}$, in Eq.~\eqref{F1} the results from both branches are element-wise added to yield the combined features${{F}_{out1}}$:
	
	\begin{equation}
		{{F}_{out1}}={{F}_{\text{b}}}+{{F}_{wt}}\label{F1}
	\end{equation}

	The combined features are then refined using a sigmoid function to generate an attention mask, which is element-wise multiplied with the global input features to apply different weighting to the input features. In Eq.~\eqref{FB} the resulting frequency-enhanced and blur perception feature map $F{{B}_{out}}$ of FEBP is defined as:
	
	\begin{equation}
		FB_{k}^{out}=F{{M}_{k}}*{{\sigma }_{sig}}\left( {{f}_{out}}\left( {{F}_{out1}} \right) \right)\label{FB}
	\end{equation}

	FEBP establishes long-range dependencies between features through strip blur perception, aggregating contextual information. The blur perception branch enhances the network's ability to deal with non-uniform blur by perceiving blur features. The frequency enhancement branch via wavelet transform addresses issues related to high-frequency information loss and poor image detail restoration during the convolution process. The combination of dual branches enhances the capability of removing non-uniform blur from blurred images.

	\subsection{Loss Function}
	
	Based on the loss function structure used in multi-scale networks ~\cite{cho2021rethinking}, we combine the multi-scale content loss and the multi-scale frequency domain reconstruction loss as follows:

	\begin{equation}
		{{L}_{total}}={{L}_{cont}}+\lambda {{L}_{MSFR}}
	\end{equation}
	where ${{L}_{total}}$ represents the total loss, ${{L}_{cont}}$ represents the content loss in Eq.~\eqref{content loss}. ${{L}_{MSFR}}$ denotes the frequency domain reconstruction loss in Eq.~\eqref{frequency loss}. And \( \lambda \) is the weight for the frequency domain reconstruction loss, set to 0.1.
	
	The multi-scale content loss function uses L1 loss, computed as:
	
	\begin{equation}
		{{L}_{cont}}=\sum\limits_{k=1}^{K}{\frac{1}{{{t}_{k}}}}{{\left\| \left. {{{I}}_{k}}-{{S}_{k}} \right\| \right.}_{1}}\label{content loss}
	\end{equation}
	where \( K = 3 \) denotes the number of network layers, ${{t}_{k}}$ represents the total number of normalized elements, ${{I}_{k}}$ is the deblurred image at scale \( k \), and ${{S}_{k}}$ is the ground truth image at scale \( k \).
	
	To reduce the discrepancy between the deblurred image and the ground truth image in terms of high-frequency information such as texture structure, a frequency domain reconstruction loss is also used. This loss is computed by first performing a 2D Fourier Transform on both the deblurred and ground truth images, and then calculating the mean absolute error between their frequency domain representations:
	
	\begin{equation}
		{{L}_{MSFR}}={{\sum\limits_{k=1}^{K}{\frac{1}{{{t}_{k}}}\left\| \left.  \mathscr{F}\left( {{{I}}_{k}} \right)-\mathscr{F}\left( {{S}_{k}} \right) \right\| \right.}}_{1}}\label{frequency loss}
	\end{equation}
	where \( \mathscr{F} \) denotes the Fast Fourier Transform (FFT).

	\section{Experiments}\label{sec4}
	
	\subsection{Experimental Setup and Dataset}
	
	The experiments are implemented on a server running the Ubuntu operating system, equipped with a GTX 4090 GPU with 24GB of memory. The network models were implemented using the PyTorch deep learning framework. The GoPro dataset was used to train the models. The GoPro dataset~\cite{nah2017deep}  includes 3,214 pairs of blurry and clear images, with 2,103 pairs used for training and 1,111 pairs for testing. Each network was trained for 3,000 epochs to ensure better convergence. Additionally, the models trained on the GoPro dataset were also tested on the HIDE dataset~\cite{shen2019human}. HIDE dataset~\cite{shen2019human} includes 8,422 pairs of blurry and clear images, divided into 6,397 pairs for training and 2,025 pairs for testing.
	 
	In this paper, we used Peak Signal-to-Noise Ratio (PSNR) and Structural Similarity Index (SSIM) as main evaluation metrics for image deblurring. PSNR is based on the Mean Square Error (MSE) definition, given a sharp image ${S}$ and a deblurred image $I$, its MSE can be defined as: 
	\begin{equation}
		\operatorname{MSE}({S}, {I}) = \frac{1}{K} \sum_{k=0}^{K-1}\left({S}_k-{I}_k\right)^2\label{mse}
	\end{equation}
	where $K$ represents the number of pixels. Then PSNR can be defined as:
	\begin{equation}
		\operatorname{PSNR}({S}, {I}) = 10 \log _{10}\left(\frac{MAX_I^2}{\operatorname{MSE({S}, {I})}}\right)\label{psnr}
	\end{equation}
	$MAX_I$ is the maximum pixel value of the image in Eq.~\eqref{psnr}. The higher the PSNR value is, the closer the deblurred image is to the sharp image. 
	
	SSIM assesses image similarity based on three characteristics: brightness, contrast, and structure. The calculation formula is as follows:
	
	\begin{equation}
		\operatorname{SSIM}({S}, {I}) = [l({S}, {I})]^\alpha \cdot [c({S}, {I})]^\beta \cdot [s({S}, {I})]^\gamma
		\label{Eq.ssim}
	\end{equation}
	where $l({I}, {S})$ represents the  comparison of brightness, $c({I}, {S})$ represents the comparison of contrast, $s({I}, {S})$ represents the comparison of structure.
	In engineering, often make $\alpha=\beta=\gamma=1$.
	In Eq.~\eqref{Eq.ssim},
	
	\begin{equation}
		l({S}, {I})= \frac{2 \mu_I \mu_S+C_1}{\mu_I^2+\mu_S^2 + C_1},~~~
		c({S}, {I}) = \frac{2 \sigma_I \sigma_S+C_2}{\sigma_I^2+\sigma_S^2 + C_2},~~~
		s({S}, {I}) = \frac{\sigma_{I S}+C_3}{\sigma_I \sigma_S + C_3},
		\label{Eq.lcs}
	\end{equation}
	
	where $\mu_I$, $\sigma_I$ and $\mu_S$, $\sigma_S$ represent the mean, standard deviation of the images $I$ and $S$.
	$\sigma_{IS} = \frac{1}{K} \sum_{k=0}^{K-1}\left({I}_k - \mu_I\right)\left({S}_k - \mu_S\right) $ indicates the covariance of the deblurred image $ I $ and the real sharp image $ S $.
	$C$ is a constant to avoid the denominator 0 causing system errors.
	A higher SSIM value indicates a greater similarity between the deblurred image and the sharp image.

	Furthermore, since PSNR and SSIM metrics still have a certain gap from human visual perception, we further evaluated the restored image quality using Learned Perceptual Image Patch Similarity(LPIPS)~\cite{zhang2018unreasonable} metric and Visual Information Fidelity (VIF)~\cite{sheikh2006image} metric on the GoPro~\cite{nah2017deep} and HIDE~\cite{shen2019human} datasets to better reflect whether the restored images align with human perception.	

	The LPIPS metric measures the perceptual similarity between restored and clear images. Compared to traditional metrics, LPIPS is more aligned with human visual perception~\cite{zhang2018unreasonable}.	
	VIF is based on natural scene statistics and human visual system models, quantifying the extent of information loss in an image. It measures the visual quality of an image by comparing the visual information between the distorted image and the reference image. VIF incorporates human visual system models, making its assessments more closely aligned with human subjective perception of image quality.
	
	\subsection{Results and Analysis}
	We compared our method MFENet with 13 existing deblurring methods: Deepblur~\cite{nah2017deep}, SRN~\cite{tao2018scale}, MSS-Net~\cite{kim2022mssnet}, MRDNet~\cite{zhang2024image}, MIMO-UNet ~\cite{cho2021rethinking}, Deblur-GAN ~\cite{kupyn2018deblurgan}, DeblurGAN-v2 ~\cite{kupyn2019deblurgan}, DBGAN~\cite{zhang2020deblurring}, MT-RNN~\cite{park2020multi}, DMPHN ~\cite{zhang2019deep}, MSCAN~\cite{wan2020deep}, SimpleNet~\cite{li2021perceptual}, and PSS-NSC~\cite{gao2019dynamic}. The comparison focused primarily on state-of-the-art (SOTA) CNN-based methods.

	\begin{table*}[t]%
		\caption{Quantitative results on the datasets GoPro and HIDE. “$\uparrow $” is denotes higher is better.\label{tab1}}	
		\begin{tabular*}{\textwidth}{@{\extracolsep\fill}lllll@{}}
			\toprule
			&\multicolumn{2}{@{}l}{\textbf{GoPro}} & \multicolumn{2}{@{}l}{\textbf{HIDE}} \\\cmidrule{2-3}\cmidrule{4-5}
			\textbf{Methods} & \textbf{PSNR(dB) $\uparrow $  }  & \textbf{SSIM $\uparrow $  }  & \textbf{PSNR(dB) $\uparrow $  }  & \textbf{SSIM $\uparrow $  }   \\
			\midrule
			DeblurGAN~\cite{kupyn2018deblurgan} &			28.27&		0.858&		24.51&		0.871\\
			Deepblur~\cite{nah2017deep}	&		29.08&	0.914&	25.73&	0.874\\
			DeblurGAN-v2~\cite{kupyn2019deblurgan} &			29.55&	0.934&	26.61&	0.875\\
			SRN~\cite{tao2018scale} 		&	30.26&	0.934&	28.36&	0.915\\
			DBGAN~\cite{zhang2020deblurring} 		&	31.10&	0.942&	28.94&	0.915\\
			MT-RNN~\cite{park2020multi}		&	31.15&	0.945&	29.15&	0.918\\
			DMPHN~\cite{zhang2019deep} 		&	31.20&	0.940&	29.09&	0.924\\
			MSCAN~\cite{wan2020deep} 		&	31.24&	0.945&	29.63&	0.920\\
			SimpleNet~\cite{li2021perceptual}	&		31.52&	0.949&	-&	-\\
			PSS-NSC~\cite{gao2019dynamic}	&		31.58&	0.948&	-&	-\\
			MIMO-UNet~\cite{cho2021rethinking}	&		31.73&	0.951&	-&	-\\
			MRDNet~\cite{zhang2024image}		&	31.79&	0.951	&29.36&	0.921\\
			MSSNet-small~\cite{kim2022mssnet} &			32.02&	0.953&	-&	-\\
			\midrule
			MFENet (Ours)&	\textbf{32.27}&	\textbf{0.956}&	\textbf{29.74}&	\textbf{0.928}\\
			
			\bottomrule
			
		\end{tabular*}
	\end{table*}

	\begin{table}[!h]%
		\caption{Results of LPIPS and NIQE on the GoPro dataset.  “$\downarrow $” is denotes lower is better. “$\uparrow $” is denotes higher is better.\label{tab2}}
		\begin{tabular*}{\textwidth}{@{\extracolsep\fill}lllll@{}}
			\toprule
			&\multicolumn{2}{@{}l}{\textbf{GoPro}} & \multicolumn{2}{@{}l}{\textbf{HIDE}} \\\cmidrule{2-3}\cmidrule{4-5}
			\textbf{Methods} & \textbf{LPIPS $\downarrow $}  & \textbf{VIF $\uparrow $}  & \textbf{LPIPS $\downarrow $}  & \textbf{VIF $\uparrow $}   \\
			\midrule
			
			DMPHN~\cite{zhang2019deep} & 0.207 &	0.9499	 &0.243   & 0.9074\\
			
			MT-RNN~\cite{park2020multi}& 0.196 &	0.9628 &	0.208 &	0.9538\\
			
			DBGAN~\cite{zhang2020deblurring}  &0.187 &	0.9570 &	0.198 & 0.9411	\\
			
			MIMO-UNet~\cite{cho2021rethinking} & 0.180 & 0.9607 &	0.198 &	0.9458\\
			
			MRDNet~\cite{zhang2024image} &0.179 &	0.9576 &	0.196 & 0.9380\\
			
			\midrule
			
			MFENet (Ours)&   \textbf{0.171}& 	\textbf{0.9689} & 	\textbf{0.193}& \textbf{0.9600}\\
			\bottomrule
			
		\end{tabular*}
		
	\end{table}

	\subsubsection{Quantitative Comparison}

	The quantitative results of the deblurring methods on the GoPro~\cite{nah2017deep} and HIDE~\cite{shen2019human} datasets are shown in Table \ref{tab1}. The results for the methods were obtained from the respective papers. 	
	As shown in Table \ref{tab1}, our proposed method achieved the best PSNR and SSIM values on both the GoPro~\cite{nah2017deep} and HIDE~\cite{shen2019human} datasets, with a PSNR of 32.27 dB and an SSIM of 0.956 on the Gopro dataset. Compared to the CNN-based benchmark model MIMO-UNet~\cite{cho2021rethinking}, MFENet improved PSNR by 0.54 dB and SSIM by 0.005 on the GoPro dataset~\cite{nah2017deep}. On the HIDE dataset~\cite{shen2019human}, compared to the latest method MRDNet~\cite{zhang2024image}, MFENet improved PSNR by 0.48 dB and SSIM by 0.005. These results demonstrate that our method MFENet effectively enhances image deblurring performance.
	
	\begin{figure*}[t]
		
		\centerline{\includegraphics[width=\textwidth]{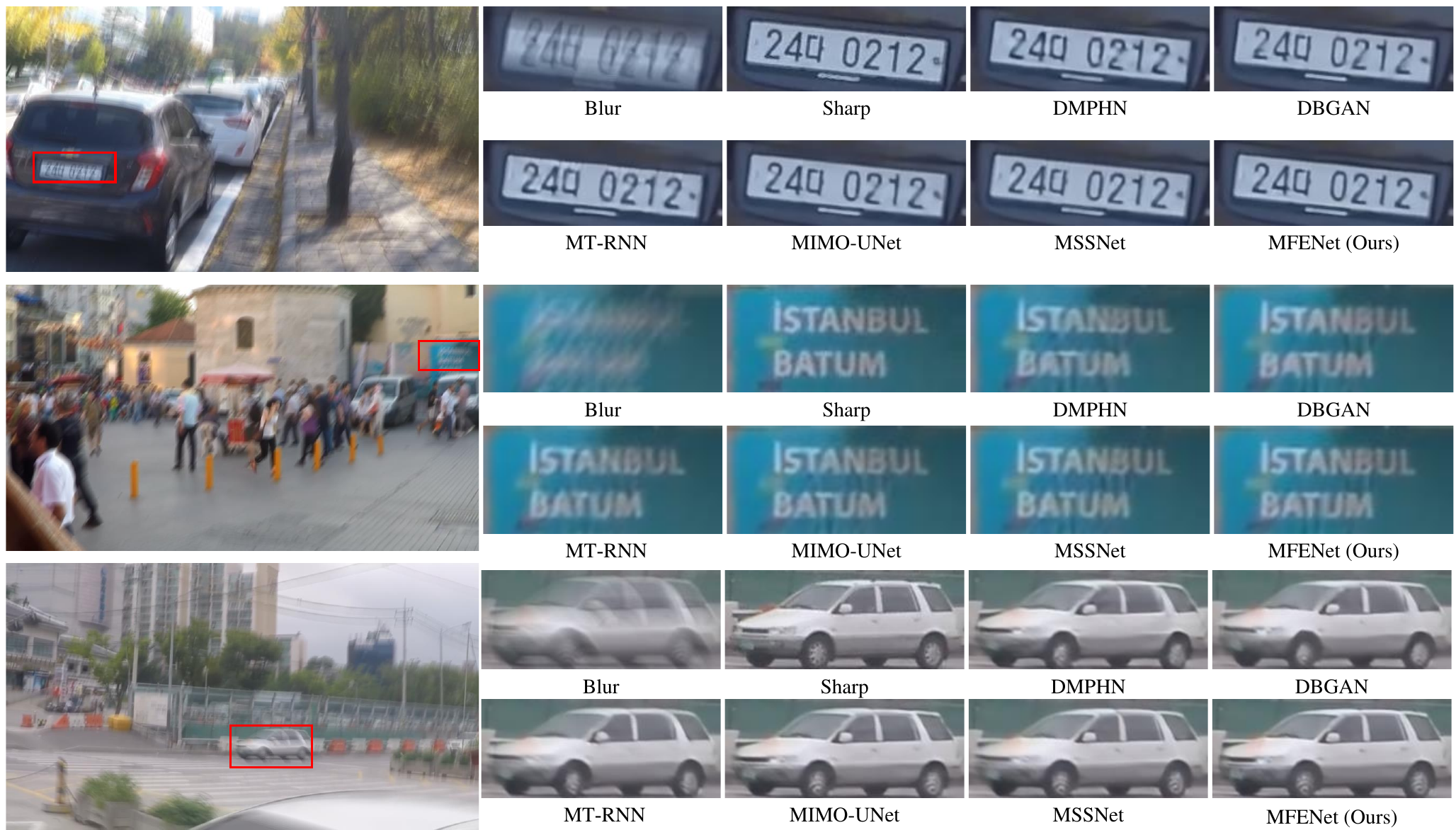}}
		\caption{Deblurring results of images from the GoPro dataset.\label{GoPro}}
	\end{figure*}
	
	A lower LPIPS score indicates that the two images are more perceptually similar, meaning the image quality is better. The experimental results are presented in Table \ref{tab2}. MFENet achieves the optimal LPIPS score compared to several CNN-based methods. Compared to the baseline model MIMO-UNet~\cite{cho2021rethinking}, our method reduced the LPIPS by 0.009 on the GoPro dataset~\cite{nah2017deep} and 	0.005 on the HIDE dataset~\cite{shen2019human}. The higher the VIF metric, the better the quality of the image. As shown in Table \ref{tab2}, MFENet achieved the highest VIF scores compared to several CNN-based methods. Compared to the baseline model MIMO-UNet~\cite{cho2021rethinking}, MFENet improved the VIF by 0.0082 on the GoPro dataset~\cite{nah2017deep}, and 0.0142 on the HIDE dataset~\cite{shen2019human}.

	\begin{figure*}[t]
		\centerline{\includegraphics[width=\textwidth]{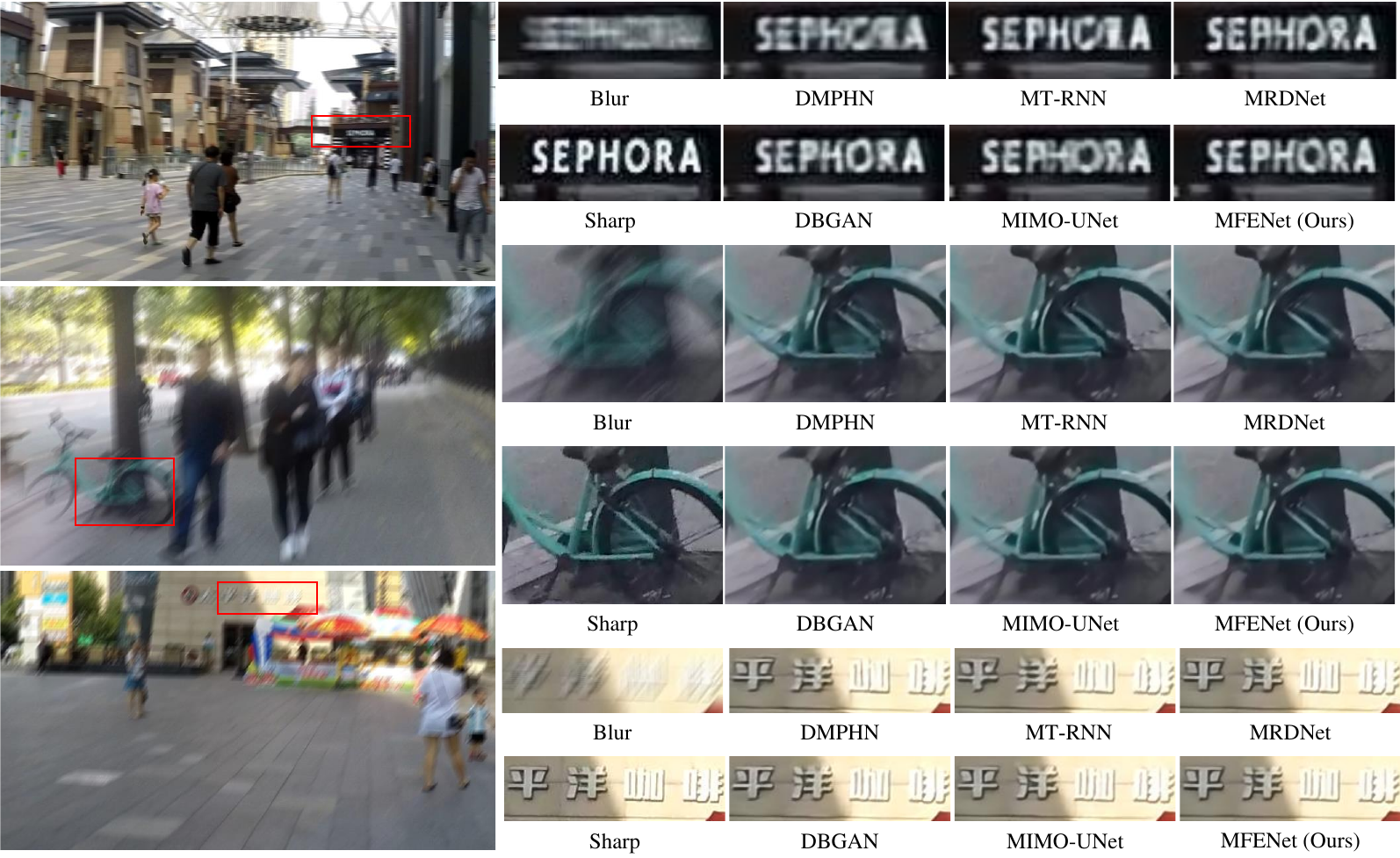}}
		\caption{Deblurring results of images from the HIDE dataset.\label{HIDE}}
	\end{figure*}

	\subsubsection{Qualitative Comparison}	
	
	To visually demonstrate the deblurring performance of our method MFENet, we compared it with several methods using the GoPro dataset~\cite{nah2017deep}, including MSSNet~\cite{kim2022mssnet}, MIMO-UNet~\cite{cho2021rethinking}, DBGAN~\cite{zhang2020deblurring}, MT-RNN~\cite{park2020multi}, and DMPHN~\cite{zhang2019deep}, as shown in Figure \ref{GoPro}. The comparison reveals that MSSNet~\cite{kim2022mssnet}, MIMO-UNet~\cite{cho2021rethinking}, DBGAN~\cite{zhang2020deblurring}, MT-RNN~\cite{park2020multi}, and DMPHN~\cite{zhang2019deep} fail to accurately capture the blurred edge information, resulting in less effective deblurring. In contrast, MFENet demonstrates superior ability to recover image texture details. Specifically, it shows the best recovery performance for elements such as license plates, text, and vehicles.

	Additionally, Figure \ref{HIDE} presents a comparison of deblurring results on the HIDE dataset~\cite{shen2019human} with MRDNet~\cite{zhang2024image}, MIMO-UNet~\cite{cho2021rethinking}, DBGAN~\cite{zhang2020deblurring}, MT-RNN~\cite{park2020multi}, and DMPHN~\cite{zhang2019deep}. The results indicate that the proposed method MFENet outperforms the other methods in recovering blurred images. However, compared to the GoPro dataset~\cite{nah2017deep}, the results of all methods on the HIDE dataset~\cite{shen2019human} are less satisfactory, highlighting a common issue with current models: limited generalization capability.

	\subsection{Ablation Study}
	
		\begin{table*}[t]%
		\centering %
		\caption{The impact of different modules on the experiments. “$\uparrow $” is denotes higher is better.\label{tab3}}%
		\begin{tabular*}{\textwidth}{@{\extracolsep\fill}lllll@{\extracolsep\fill}}
			\toprule
			\textbf{Method} & \textbf{PSNR(dB) $\uparrow $}  & \textbf{SSIM $\uparrow $} \\
			\midrule
			
			Baseline&	31.46	&0.948\\
			Baseline+ MS-FE	&31.62&	0.949\\
			Baseline+ FEBP&	31.67&	0.950\\
			Baseline+ MS-FE + FEBP	& 31.76	& 0.951\\
			Baseline+ MS-FE + FEBP + 20 Resblocks (Ours) & \textbf{32.27}	& \textbf{0.956}\\
			\bottomrule
		\end{tabular*}
	\end{table*}
	
	\begin{figure*}[t]
		\centerline{\includegraphics[width=\textwidth ]{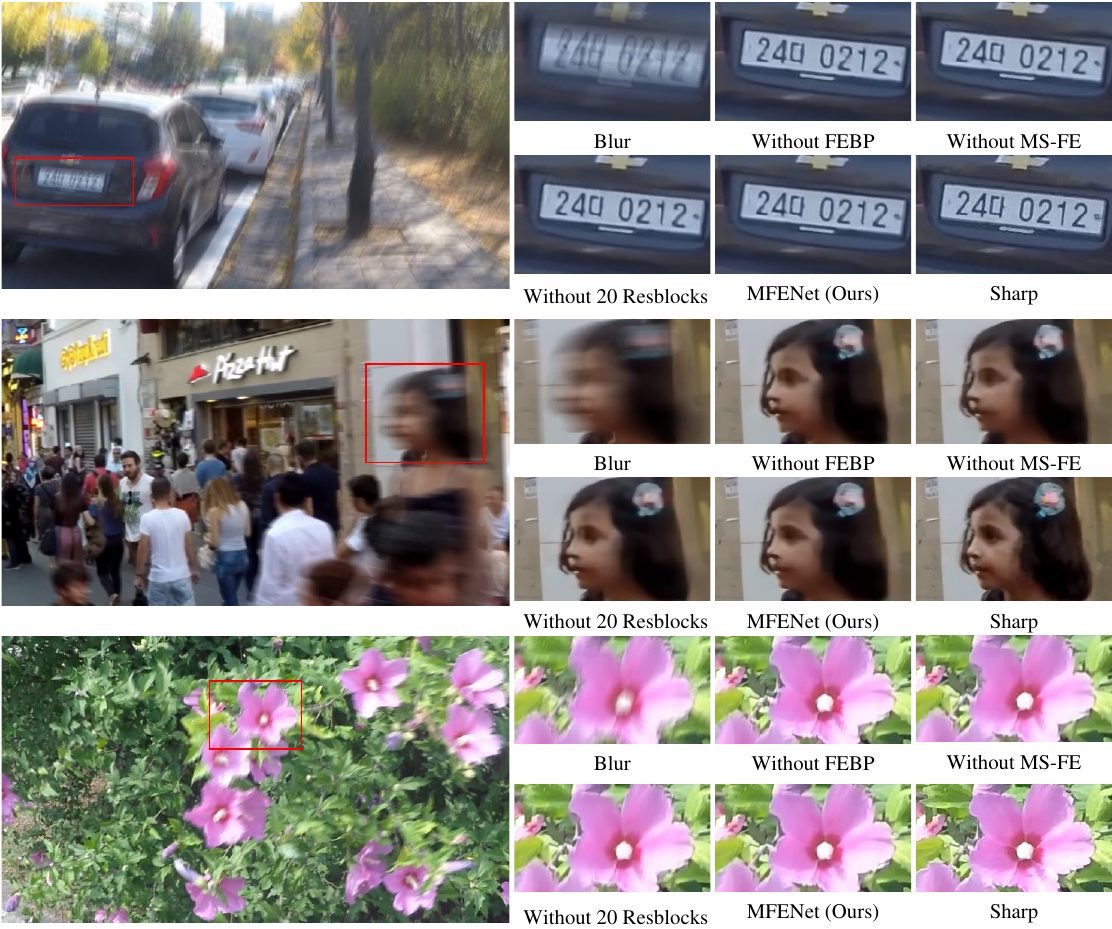}}
		\caption{The effect of different components in our model.\label{ablation}}
	\end{figure*}

	To validate the effectiveness of the Multi-scale Feature Extraction Module (MS-FE) and the Frequency Enhanced Blur Perception Module (FEBP) in the proposed method, we conducted an ablation study. Table \ref{tab3} summarizes the impact of these modules on model performance.
	
	As shown in Table \ref{tab3}, both individual modules improve performance metrics compared to the baseline network model. Adding the MS-FE module alone results in a PSNR improvement of 0.16 dB and an SSIM increase of 0.001. Similarly, incorporating the FEBP module alone yields a PSNR improvement of 0.21 dB and an SSIM increase of 0.002. When both modules are included together, the network architecture improves PSNR by 0.3 dB and SSIM by 0.003 compared to the baseline model. Specifically, When the number of network residual blocks is increased to 20, the performance of the network is significantly improved. The combined network MFENet improves PSNR by 0.81 dB and SSIM by 0.008 compared to the baseline model. The visualization results of the partial ablation experiment are shown in Figure \ref{ablation}.
	
	These results demonstrate that both the MS-FE and FEBP contribute significantly to better deblurring performance, thereby improving the recovery of image detail.
	
	\subsection{Object Detection Performance Evaluation}
	
		\begin{figure}[t]
		\centerline{\includegraphics[width=10cm]{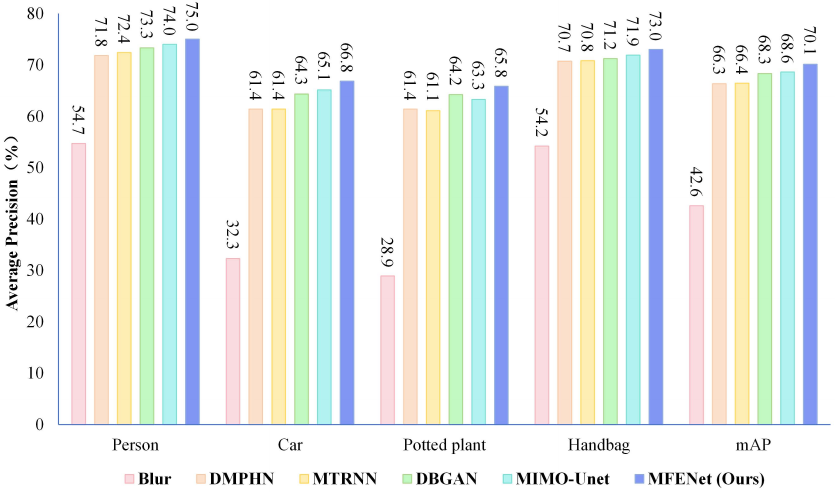}}
		\caption{Bar chart for object detection performance.\label{band}}
	\end{figure}
	
	As a preprocessing technique, image deblurring algorithms can impact subsequent computer vision tasks. In object detection, the loss of object edge contours and the presence of motion blur artifacts in blurred images make it challenging for detection algorithms to accurately recognize and locate objects. Deblurring blurred images can enhance object clarity, thereby improving the accuracy of object detection. In our experiments, we applied the Yolo-v8 object detection algorithm to images restored by MIMO-UNet ~\cite{cho2021rethinking}, DBGAN~\cite{zhang2020deblurring}, MT-RNN~\cite{park2020multi}, DMPHN ~\cite{zhang2019deep}, and the proposed method MFENet.

	\subsubsection{Quantitative Comparison}

	\begin{table*}[!t]%
		\centering %
		\caption{Comparison table of Average Percision for object detection.\label{tab4}}%
		
		\scalebox{0.6}{
			\begin{tabular}{lllllll}
				\toprule
				\textbf{Average Percision ($\%$)}& \textbf{Blur} & \textbf{DMPHN}~\cite{zhang2019deep}&	\textbf{MTRNN}~\cite{park2020multi} &  \textbf{DBGAN}~\cite{zhang2020deblurring} & \textbf{	MIMO-UNet}~\cite{cho2021rethinking} & \textbf{MFENet (Ours)}   \\
				\midrule
				Person&	54.7&	71.8&	72.4&	73.3&	74.0&	\textbf{75.0}\\
				Car&	32.7&	61.4&	61.4&	64.3&	65.1&	\textbf{66.8}\\
				Potted plant	&28.9&	61.4&	61.1	&64.2&	63.3	& \textbf{65.8}\\
				Handbag	&54.2&	70.7&	70.8&	71.2&	71.9&	\textbf{73.0}\\
				mAP&	42.6&	66.3	&66.4	&68.3&	68.6&	\textbf{70.1}\\
				
				\bottomrule
			\end{tabular}
		}
	\end{table*}

	Considering that some images in the blurred image dataset are from scenes such as streets or buildings and do not contain detectable objects, we used 634 images from the GoPro dataset~\cite{nah2017deep} containing person, car, potted plants, and handbags for object detection. We measured the Average Precision (AP) for each object class and the mean Average Precision (mAP) for performance evaluation. As shown in Figure \ref{band} and Table \ref{tab4}, MFENet achieved the highest average precision across several categories and in overall mAP compared to other methods. Specifically, MFENet shows a 20.3$\%$ improvement in detection precision for the Person category, a 34.1$\%$ improvement for the Car category, a 36.9$\%$ improvement for the Potted Plant category, and an 18.8$\%$ improvement for the Handbag category, resulting in a total average detection precision increase of 27.5$\%$.
	
	\begin{figure*}[t]
	\centerline{\includegraphics[width=\textwidth]{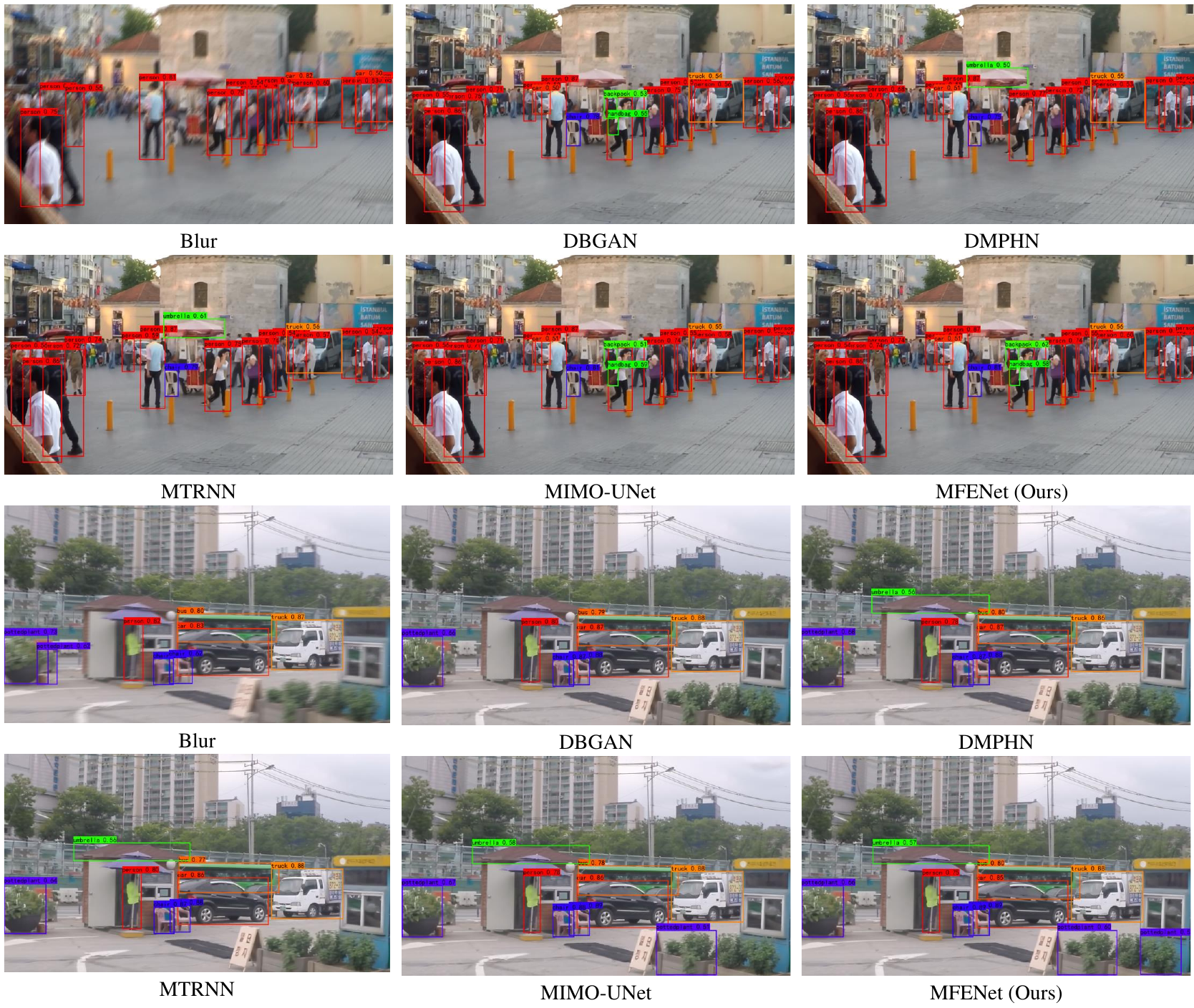}}
	\caption{Examples of object detection results.\label{detection}}
\end{figure*}
	
	\subsubsection{Qualitative Comparison}
	
	Figure \ref{detection} illustrates some of the object detection results. From the comparisons in the first and second rows, it is evident that for the Person category, MFENet significantly reduces the missed detection rate compared to blurred images and enhances the confidence in Person detection compared to other methods. For the Handbag category, MFENet accurately detects Handbags, while blurred images and deblurring methods MT-RNN~\cite{park2020multi} and DMPHN~\cite{zhang2019deep} show missed detections. Compared to DBGAN~\cite{zhang2020deblurring} and MIMO-UNet~\cite{cho2021rethinking}, MFENet improves confidence in detecting Handbags.
	
	In the comparisons in the third and fourth rows, MFENet can accurately identify the Potted Plant category compared to blurred images and other methods, which all exhibit some missed detections. Although there are fewer Car targets in the images, MFENet shows a slight improvement in recognition confidence across the three types of car compared to other methods.

	\section{Conclusions}\label{sec5}
	
	This paper presents an end-to-end multi-scale image deblurring network MFENet based on multi-scale feature extraction and frequency-enhanced blur perception. We designed a multi-scale feature extraction module (MS-FE) using depthwise separable convolutions, which effectively extracts multi-scale feature information from images, expands the receptive field, and enhances the network's performance in removing non-uniform blur. Additionally, we introduced a frequency enhanced blur perception module (FEBP) that uses multi-strip pooling to establish long-range contextual dependencies, improving the perception of blurred features in images. Moreover, wavelet transform is utilized to enhance the recovery of edge and texture details in images. Experimental results demonstrate that our method MFENet effectively deblurs images and addresses the issue of high-frequency information loss during image restoration.

\section*{Declarations}

\begin{itemize}
\item Funding:This research received no external funding.
\item Conflict of interest disclosure: The authors report there are no competing interests 
to declare.
\item Author contribution:	Methodology, Y.X.; 
Supervision, Z.L. and Y.X.; 
Writing—Original Draft, Y.X., H.Z. and C.L.; 
Writing—Review and Editing, Y.X.and H.Z.;
Visualization, Y.X., C.L. and H.Z.;
All authors have read and agreed to the published version of the manuscript.
\end{itemize}

\bibliography{sn-bibliography}
\balance

\end{document}